\documentclass[letterpaper]{article} % DO NOT CHANGE THIS
\usepackage{aaai25}  % DO NOT CHANGE THIS
\usepackage{times}  % DO NOT CHANGE THIS
\usepackage{helvet}  % DO NOT CHANGE THIS
\usepackage{courier}  % DO NOT CHANGE THIS
\usepackage[hyphens]{url}  % DO NOT CHANGE THIS
\usepackage{graphicx} % DO NOT CHANGE THIS
\usepackage{natbib}  % DO NOT CHANGE THIS AND DO NOT ADD ANY OPTIONS TO IT
\usepackage{caption} % DO NOT CHANGE THIS AND DO NOT ADD ANY OPTIONS TO IT
\usepackage{algorithm}
\usepackage{algorithmic}
\usepackage{amsmath}
\usepackage{multirow}
\usepackage{amssymb}
\usepackage{booktabs}
\usepackage{cleveref}
\usepackage{newfloat}
\usepackage{listings}
\DeclareCaptionStyle{ruled}{labelfont=normalfont,labelsep=colon,strut=off} % DO NOT CHANGE THIS
\floatstyle{ruled}
\newfloat{listing}{tb}{lst}{}
\floatname{listing}{Listing}
\title{Planning from Imagination: Episodic Simulation and\\Episodic Memory for Vision-and-Language Navigation}
\author{
    Yiyuan Pan,
    Yunzhe Xu,
    Zhe Liu\thanks{Corresponding Author. Yiyuan Pan and Hesheng Wang are with Department of Automation, Shanghai Jiao Tong University, China. Yunzhe Xu and Zhe Liu are with MoE Key Lab of Artificial Intelligence, AI Institute, Shanghai Jiao Tong University, China.},
    Hesheng Wang
    }
\affiliations {
    Shanghai Jiao Tong University\\
    \{pyy030406, xyz9911, liuzhesjtu, wanghesheng\}@sjtu.edu.cn
}

\usepackage{bibentry}

\begin{document}

\maketitle

\begin{abstract}
Humans navigate unfamiliar environments using episodic simulation and episodic memory, which facilitate a deeper understanding of the complex relationships between environments and objects. Developing an imaginative memory system inspired by human mechanisms can enhance the navigation performance of embodied agents in unseen environments. However, existing Vision-and-Language Navigation (VLN) agents lack a memory mechanism of this kind. To address this, we propose a novel architecture that equips agents with a reality-imagination hybrid memory system. This system enables agents to maintain and expand their memory through both imaginative mechanisms and navigation actions. Additionally, we design tailored pre-training tasks to develop the agent's imaginative capabilities. Our agent can imagine high-fidelity RGB images for future scenes, achieving state-of-the-art result in Success rate weighted by Path Length (\texttt{SPL}).
\end{abstract}

% Uncomment the following to link to your code, datasets, an extended version, or similar.

\section{Introduction}

Autonomous embodied agent navigation is advancing through Vision-and-Language Navigation (VLN) research \cite{VLN_R2R, VLN_RVR}. In VLN tasks, an agent needs to reach the target location given navigation instructions. This task, however, is more challenging when agents navigate in unseen environments, with performance degradation compared to seen environments.

\begin{figure}[!t]
\centering
\includegraphics[width=0.47\textwidth]{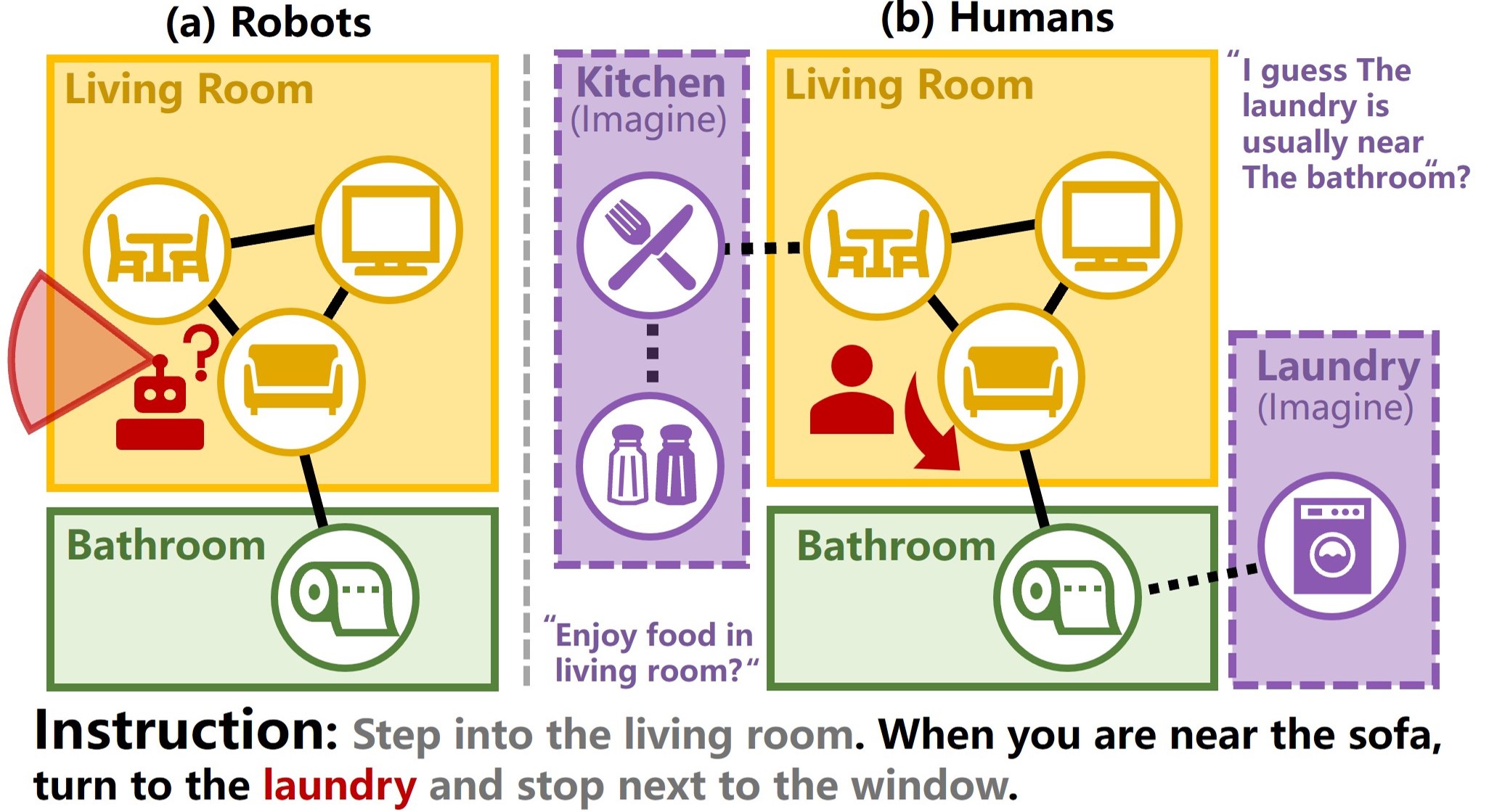} % Reduce the figure size so that it is slightly narrower than the column.
\caption{Humans utilize episodic memory and episodic simulation to recall past experiences and predict future outcomes in unfamiliar environments. In contrast, navigation agents often struggle in unseen environments due to their inability to construct and leverage such cognitive frameworks.}
\label{fig1}
\end{figure}

Human navigation behaviors in unfamiliar environments provide valuable insights, particularly through the neuroscience concepts of episodic simulation and memory \cite{Episodic1, Episodic2}, which may help address this challenge. These mechanisms enable humans to use memory to mentally simulate uncertain information or scenes, aiding decision-making. Humans can imagine fine-grained visual features (e.g., RGB images) and higher-level spatial structures (e.g., location distributions) in unseen environments \cite{Episodic3}. As illustrated in \Cref{fig1}, given an instruction like ``go to the laundry room", humans might imagine a room with washing machines, likely adjacent to a bathroom encountered earlier, using such mental simulations to infer navigation directions.

While existing approaches have incorporated mechanisms for imagining unseen environments, they lack the ability to integrate the imagined results into time-series persistent memory. Specifically, This prevents the formation of long-term episodic simulations, which is essential for effective navigation \cite{Imagine1, Imagine4}. Current methods enable agents to predict potential object locations \cite{Pos2} and RGB features \cite{Obj1, Obj2}, but these predictions are transient and overwritten at each step, precluding durable memory representations. Such limitations hinder agents from fusing visual and contextual information to support future reasoning or decision-making. For instance, in \Cref{fig1}, an agent's inability to infer the presence of spice bottles from contextual cues such as a knife and fork illustrates the shortcomings of existing mechanisms. Consequently, these methods produce isolated and short-lived imaginative outputs, lacking the continuity and compositional depth characteristic of human episodic memory and simulation.

In this paper, we introduce the \textbf{S}pace-\textbf{A}ware \textbf{L}ong-term \textbf{I}maginer (SALI) agent, the first navigation agent explicitly designed to emulate human-like episodic simulation and memory. SALI leverages an imaginative memory system to capture both high-level spatial structures of the environment and fine-grained RGB features. The generation of these imaginative outputs is achieved by integrating prior imagined results with historical information, forming a recurrent imagination module. This iterative process enables SALI to maintain a hybrid memory system that combines real and imagined representations, facilitating robust reasoning and navigation in complex and unseen environments. To implement this system, we model the agent’s memory using a topological map \cite{Map_Memory}, where nodes represent spatial locations enriched with RGB-D and semantic features generated by the imagination module or the navigation actions. Multimodal transformers then encode these nodes alongside natural language instructions, enabling reasonable navigation decisions at each step. The integration of imaginative and realistic information allows SALI to reason holistically about its environment while dynamically adapting to new scenarios. As a result, SALI achieves state-of-the-art (SoTA) performance on R2R \cite{VLN_R2R} and REVERIE \cite{VLN_RVR}, highlighting its effectiveness.

To summarize, our contributions are as follows.

\begin{itemize}
    \item Our proposed SALI has human-like episodic memory and episodic simulation abilities, enhancing general navigation performance. Viewing from the perspective of human brains, we endow the agent with the ability to learn anticipatory knowledge through imagination.
    \item We established a recurrent, end-to-end imagination module that generates high-fidelity RGB scene representations. SALI dynamically fuses imagined scenes with real observations into a hybrid memory map, ensuring effective navigation decisions.
    \item SALI achieves SoTA performance on R2R and REVERIE, improving \texttt{SPL} by 8\% and 4\% in unseen scenarios, respectively.
\end{itemize}

\section{Related Works}

\noindent\textbf{Vision-and-language Navigation. } Guiding robot navigation using visual inputs and natural language instructions is a core task in embodied AI \cite{VLN_R2R, VLN_RVR}. A key challenge lies in effectively aligning multimodal visual-linguistic inputs \cite{Multimodal}. Existing VLN approaches \cite{ViT, Obj2, Recurrent} often falter in unseen environments, struggling to associate unfamiliar visual observations, such as novel materials and textures, with navigation instructions, leading to degraded performance. To address these limitations, we propose a novel VLN architecture inspired by human episodic simulation and memory, enhancing robustness and performance in unseen environments.

\noindent\textbf{Memory Mechanism. } Constructing memory for long-term decision support is critical for embodied agents. SLAM has been widely used to build memory maps in navigation \cite{SLAM1, SLAM2}. To reduce memory overhead, more methods such as 2D bird-eye-view maps \cite{Imagine4, BEV} or ego-centric grid maps \cite{Egocentric, GridMM} are investigated to help agents better understand spatial information. Additionally, some other navigation agents employ topological maps to further simplify and extract high-level features to learn underlying spatial knowledge \cite{Imagine4}. However, Integrating historical memory with future imagination is underexplored, limiting navigation robustness in unseen environments. Drawing inspiration from human cognition, we design an episodic simulation-based memory mechanism to enhance agents' foresight.

\noindent\textbf{Prediction during Navigation. } Inspired by video prediction \cite{Predict2}, prior works have adopted imaginary mechanisms to support navigation decisions by generating fine-grained visual information. They use RGB-D images to generate high-fidelity images \cite{Obj1, Imagine1}. Furthermore, there is also pioneering work that explores the generation of spatial structural information \cite{Imagine2, Imagine3} with the help of depth cameras \cite{Pos1, Ladar}. However, xisting imagination mechanisms operate in isolation, without integration into persistent memory. We propose a memory-based framework to bridge transient predictions with long-term memory, enabling reusable navigation information. Our adaptive mechanism integrates imagination into memory for durable, context-aware navigation.

\begin{figure*}[!t]
\centering
\includegraphics[width=0.95\textwidth]{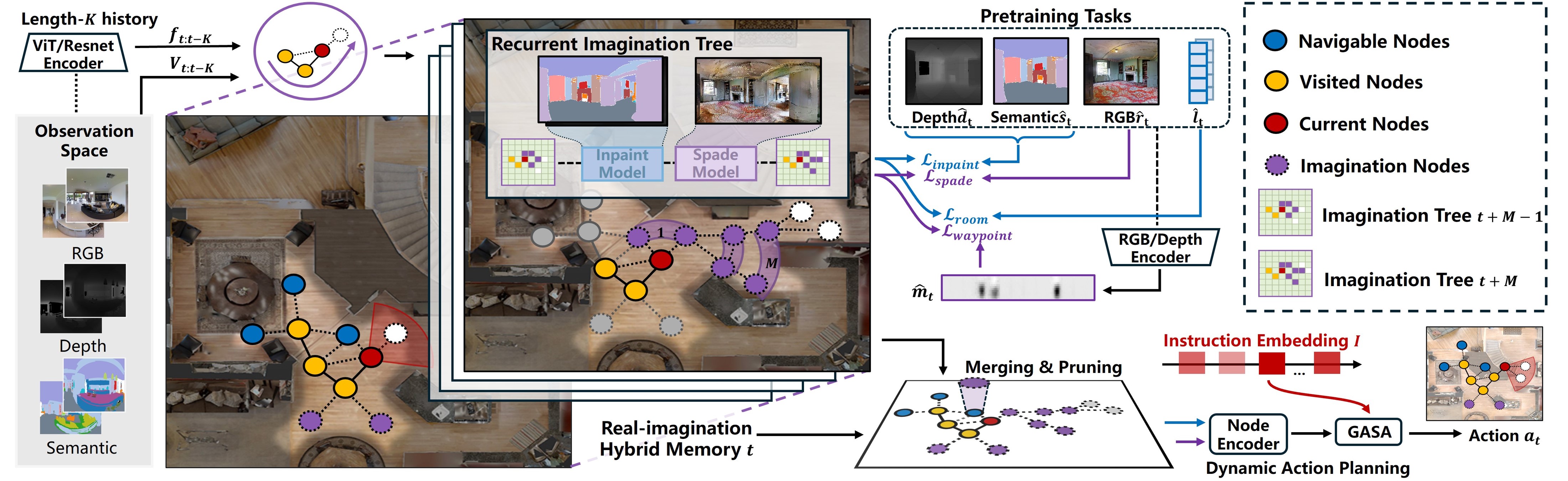} % Reduce the figure size so that it is slightly narrower than the column.
\caption{We propose building a hybrid imagination-reality memory for long-term navigation decisions. Based on the navigation observation and trajectories, the agent will imagine future scenes for unvisited environments. The imagination will then be fused into its hybrid memory to aid further decision-making. The figure also illustrates a series of pre-training tasks that we propose for the imagination module.}
\label{fig2}
\end{figure*}

\section{Method}
\textbf{Problem Formulation. } The VLN task is set in a discrete environment $\mathcal{G} = \{\mathcal{V}, \mathcal{E}\}$ \cite{VLN_R2R}, where $\mathcal{V}$ is navigation nodes and $\mathcal{E}$ represents the connectivity relations. The VLN dataset provides instruction set $I$ and corresponding ground-truth navigation paths. It requires guiding a robot to accomplish navigation tasks with given instructions. At each navigation step $t$, the agent can acquire the RGB image $r_t$ and depth image $d_t$ via the camera and the current position information $p_t$ via the GPS sensor. Following \cite{Semantic}, an agents is equipped with a visual classifier to acquire semantic images $s_t$. The agent needs to learn a policy to make actions based on observations $O_t= \{r_t,d_t,s_t,p_t\}$ and instruction.

\noindent \textbf{Method Overview. } As shown in \Cref{fig2}, SALI has a human-like episodic simulation and episodic memory mechanism. At each navigation step $t$, the agent will maintain a topological map as its memory to store both realistic and imaginative information and make navigation decisions based solely on the memory (Section 3.1). Then, the agent will use its memory to imagine future information of both high-level spatial knowledge and low-level image features and merge the imaginary into the memory (Section 3.2). We conclude this part by presenting our approach to training and inference in Section 3.3.

\subsection{Real-Imaginary Hybrid Memory} \label{Real-Imaginary Hybrid Memory}
To provide the agent with the ability of global action planning, we construct a mixed-granularity memory. As shown in \Cref{fig2}, at navigation step $t$, the agent will update the topological memory $G_t$ based on the current observation $O_t$ and historical information.

\subsubsection{3.1.1 Memory Map Representation}
The topological map memory is represented as $G_t = \{N_t, E_t\}$, where $E_t$ records the Euclidean distance between neighboring nodes, and $N_t$ represents the nodes containing visual inputs $V_t$, position information $p_t$, and node feature $f_t$. Visual inputs $V_t = \left\{ r_t,d_t,s_t \right\}$, stored in equirectangular panorama form, is used to generate fine-grained images. We use a pre-trained Vision Transformer (ViT) \cite{ViT} and ResNet encoder to get the node feature $f_t$ from visual inputs $V_t$.

Nodes $N_t$ can be classified into four categories: visited nodes \includegraphics[width=0.02\textwidth]{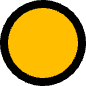}, current nodes \includegraphics[width=0.02\textwidth]{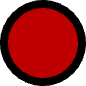}, navigable nodes \includegraphics[width=0.02\textwidth]{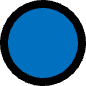}, and imagination nodes \includegraphics[width=0.02\textwidth]{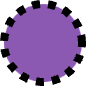}. Each node type stores different information. For $V_t$, visited and current nodes store complete and fixed image information, while the navigable node's $V_t$ is represented by partial visual inputs observed from neighboring real nodes and contains potential information from imagination nodes. The $V_t$ of the imagination node is completely generated from imagined images. For $p_t$, the agent can only reach the potential position associated with navigable nodes, as locations corresponding to \includegraphics[width=0.02\textwidth]{Images/imagination_node.png}, \includegraphics[width=0.02\textwidth]{Images/current_node.png}, or \includegraphics[width=0.02\textwidth]{Images/visited_node.png} are not accessible via action $a_t$.

To manage memory efficiency, pruning operations are applied to newly imagined nodes. If two nodes are determined to represent the same node, the pruning operation retains a single node, and its feature $f_t$ is updated using the average pooling of the features from both nodes. The new node remains an imagination node when both are imagination nodes, otherwise is labeled as a navigable node. In addition, we set an upper bound of $\bar{N}$ on the number of imagination nodes. The criteria of pruning are based on the feature cosine similarity and the negative position mean square error (MSE) between the imagination node $N_i$ and other node $N_j$:
\begin{align}
    \text{Criterion}(N_i,N_j) =\frac{f_if_j}{||f_i|| ||f_j||}&-\text{MSE}(p_i, p_j).
\end{align}

\begin{figure*}[!t]
\centering
\includegraphics[width=0.95\textwidth]{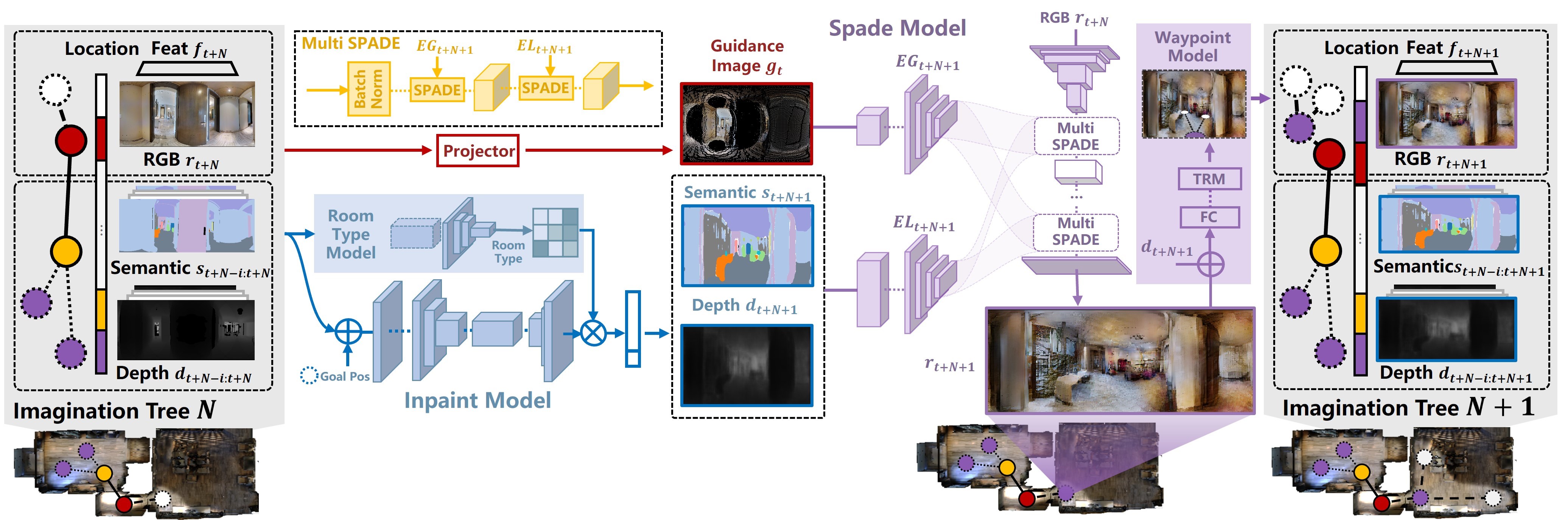} % Reduce the figure size so that it is slightly narrower than the column.
\caption{The imagination model includes four pre-trained models (inpaint model, spade model, room-type model, and waypoint model). We propose an end-to-end architecture that allows local imaginary trees to continuously update and maintain themselves, producing high-quality depth images, semantic images, and RGB images.}
\label{fig3}
\end{figure*}

\subsubsection{3.1.2 Dynamic Action Planning} 
As shown in \Cref{fig2}, the topological memory $G_t$ at time $t$ is processed by a multimodal transformer to obtain contextual representations. this transformer contains pre-trained encoders for real nodes ( \includegraphics[width=0.02\textwidth]{Images/visited_node.png}, \includegraphics[width=0.02\textwidth]{Images/navigable_node.png}, \includegraphics[width=0.02\textwidth]{Images/current_node.png} ), fine-tuned encoders for imagination nodes ( \includegraphics[width=0.02\textwidth]{Images/imagination_node.png} ), and a pre-trained instruction encoder. The real node encoder and the imagination node encoder share the same structure, unified as the node encoder for simplicity.

\noindent\textbf{Node Embedding. } The input to generate node embeddings $\hat{V}_t$ includes three elements: node feature $f_t$, location encoding, and navigation step encoding. The latter marks visited nodes with their last visited time ($0$ for \includegraphics[width=0.02\textwidth]{Images/imagination_node.png}, \includegraphics[width=0.02\textwidth]{Images/navigable_node.png}). This encoding reflects the temporal structure of memory. Node embeddings $\hat{V}_t$ are categorized into $\hat{V}_t^r$ and $\hat{V}_t^i$ based on whether the nodes are real or imagined. Furthermore, a `stop' node is added to the memory to represent a stop action and is connected to all other nodes.

\noindent\textbf{Instruction and Node Encoders. } Each word embedding in instruction $I$ is augmented with a positional embedding and a type embedding. A multi-layer transformer processes these tokens to generate contextual representations, denoted as instruction embeddings $\hat{I}$.

Node and word embeddings $\hat{V}_t$ and $\hat{I}$ are fed into a multi-layer cross-modal transformer. Following \cite{Map_Memory}, we endow the transformer with a graph-aware self-attention (GASA) layer to capture the environment layout. The network outputs navigation scores $s_i^r$ and $s_i^i$ for each navigable and imagination node as follows, where FFN is a two-layer feed-forward network:
\begin{align}
    s_t^k = \text{FFN}(\text{GASA}(\hat{V}_t^k)), k=i,r.
\end{align}

\noindent\textbf{Action Fusion Policy. } We propose an adaptive framework for navigation decision-making, which involves the dynamic integration of navigation scores obtained from both navigable nodes and imagination nodes. Initially, navigation scores from imagination nodes will be multiplied by a fusion factor $\gamma_t$ and added to the scores of the nearest navigable nodes in Euclidean distance, since they are not reachable. The fusion factor is generated by concatenating the imagination nodes and real nodes and then feeding it into an FFN layer. The scores for next-node selection are obtained by:
\begin{align}
    \gamma_t = \text{Sigmoid}(\text{FFN}(\left[\hat{V}^r_t, \hat{V}^i_t\right])),  \\
    \hat{s_t} = s_t^r+\sum\nolimits _{s^i_{t} \in \mathcal{S}(i)}  \gamma_t s^i_{t},
\end{align}
where $\mathcal{S}(i)$ represents the set of all imagination nodes that need to be added for navigable node $N_i$.

\subsection{Recurrent Imagination Tree}
We put forth a recurrent imagination module with a tree structure. Drawing inspiration from \cite{Predict1, Predict2}, SALI can generate high-resolution future images. At step $t$, a length-$K$ history information $H_t = \{d_{t-K:t},s_{t-K:t},p_{t-K:t}\}$, the RGB image $r_t$, and the neighboring position of the current position $p^g_t$---all extracted from navigation memory $G_t$---are utilized to initialize imaginary tree $T_t^0=\{H_t,r_t,p^g_t\}$. The imaginary tree $T_t^M$ is then generated iteratively and finally integrated with the $G_t$ as mentioned in Section 3.1.1.

\subsubsection{3.2.1 Recurrent Imagination Mechanism}
The imagination tree grows iteratively. Without loss of generality, \Cref{fig3} illustrates the process of expanding a recurrent imagination tree of imagination step $N+1$ ($N+1 \le M$) from step $t+N$. We first use the input queue $T^{N}_t$ to generate a list of structured label images $s^{N+1}_t$ and $d_t^{N+1}$ at imagination step $N+1$ by the inpaint model for each navigable position $p^{g, N}_{t}$. Subsequently, the guidance image $g^{N}_t$ is generated through a point cloud projector. Finally, the RGB image $r_{t}^{N+1}$ and the target position $p^{g, N+1}_{t}$ will be generated by the spade model with the guidance image $g_t^{N}$, the semantic image $s_t^{N+1}$, and the depth image $d_t^{N+1}$ as inputs. The short-term memory is then updated to the state $T^{N+1}_t$. It is noteworthy that all images are uniformly oriented with $0$ headings to prevent the generation of different images for the same point due to variations in agent action orientations.

\subsubsection{3.2.2 High-fidelity Image Generation}
\

\indent \textit{1) Inpaint Model. } The inpaint model is an encoder-decoder-based network designed for depth and semantic image generation. For each navigable position $p^{g, N}_t$, the process begins by converting historical data $H_t$ into corresponding structural maps $x_t^{N+1}$ through a point cloud projector. Subsequently, the map is fed into an encoder-decoder based on RedNet and ResNet \cite{Resnet} to generate one-hot semantic code $e_t^{N+1}$ and $d_t^{N+1}$, obtained by:
\begin{align}
    X_t &= \text{Encoder-Decoder}(x_t), \\
    [d_{t+1},e_{t+1}] &= \sigma C_S(X)+(1-\sigma)C_F(X_t),
\end{align}
where $\sigma$ is denoted as weight parameter $C_G(X_t)$. $C_G,C_S,C_F$ stand for Gate-, Shortcut- and Final-Convolution layer.

\begin{figure}[!t]
\centering
\includegraphics[width=0.45\textwidth]{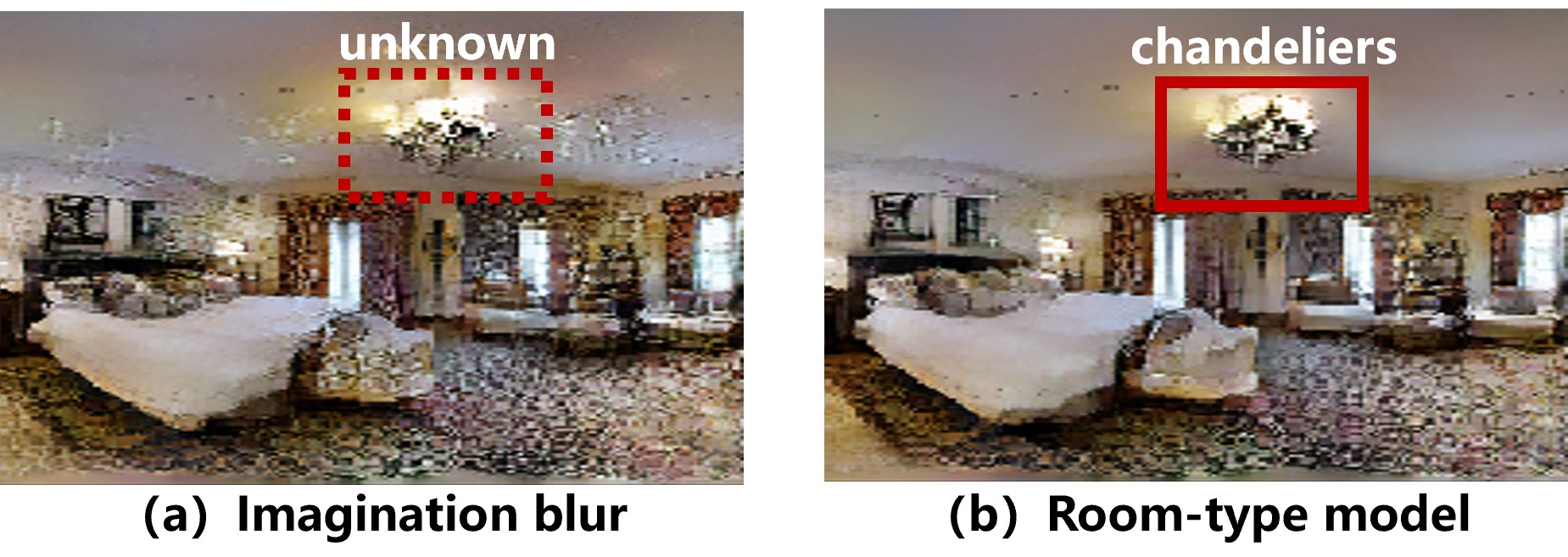} % Reduce the figure size so that it is slightly narrower than the column.
\caption{Before and after adding room-type model.}
\label{fig4}
\end{figure}

As the imagination step upper bound $M$ increases, the generated semantic images become increasingly ambiguous, with overlapping likelihoods for objects in the same pixel (\Cref{fig4}) \cite{Blur}. The likelihood of different objects in the same pixel would be similar. For this reason, we incorporate the room type model. Given the high correlation between room information and objects (e.g., refrigerator-kitchen, bed-bedroom), SALI can ascertain the presence of specific objects by determining the room type based on commonsense knowledge. SALI uses a predefined object weight dictionary $w$ for each room, refining one-hot semantic code $e_{t+1}$ as $e_{t+1}\cdot(1+w)$. The updated code is then transformed to generate $s_{t+1}$.

\textit{2) Spade Model. } The spade model is an image-to-image translation GAN \cite{GAN1, GAN2} that enables high-resolution RGB generation capability \cite{Spade}. The input to the spade model includes the embeddings of the RGB image $r_t^{N}$ and the guidance image $g_t^{N}$, and the output is the RGB image $r_t^{N+1}$.

For the following navigable position prediction $p_t^{g, N+1}$, the generated RGB-D images $r_t^{N+1}$ and $d_t^{N+1}$ are fed into a BERT-based \cite{2019bert} waypoint model.

\subsubsection{3.2.3 Cross-correction}
Global memory and local imagination mechanisms are designed to complement each other. Imagination enhances the image features of navigable nodes and facilitates map expansion. Conversely, the global map provides historical context and trajectory options, reducing information loss for imagination processes.

\subsection{Training and Inference}
\subsubsection{3.3.1 Multimodal Transformer Pre-training}
Previous studies have shown the benefits of pre-training VLN models with auxiliary tasks. We pre-train our model using expert behavior and imitation learning. Pre-training tasks include masked language modeling (MLM) \cite{2019bert}, masked region classification (MRC) \cite{ViT}, single-step action prediction (SAP) \cite{Pre1}, and object grounding (OG) \cite{OG}.

\begin{table*}[t]
\centering
\fontsize{9pt}{11pt}\selectfont
\setlength{\tabcolsep}{1mm}
\begin{tabular}{l cccc | cccc || ccc | ccc}
\toprule
 & \multicolumn{4}{c}{Val Unseen-R2R} & \multicolumn{4}{c}{Test Unseen-R2R} & \multicolumn{3}{c}{Val Unseen-REVERIE} & \multicolumn{3}{c}{Test Unseen-REVERIE} \\
\cmidrule(r){2-5} \cmidrule(r){6-9} \cmidrule(r){10-12} \cmidrule(r){13-15}
\textbf{Methods} & 
\textbf{NE}$\downarrow$ & OSR$\uparrow$ & \textbf{SR}$\uparrow$ & \textbf{SPL}$\uparrow$ & 
\textbf{NE}$\downarrow$ & OSR$\uparrow$ & \textbf{SR}$\uparrow$ & \textbf{SPL}$\uparrow$ &
\textbf{SR}$\uparrow$ & \textbf{RGS}$\uparrow$ & \textbf{RGSPL}$\uparrow$ &
\textbf{SR}$\uparrow$ & \textbf{RGS}$\uparrow$ & \textbf{RGSPL}$\uparrow$ \\
\midrule
Seq2Seq \cite{VLN_R2R}
& 7.81 & 28 & 21 & - 
& 7.85 & 27 & 20 & - &
56 & 36 & 26 
& 55 & 32 & 22 \\
RecBert \cite{Recurrent}
& 3.93 & - & 63 & 57 
& 4.09 & 70 & 63 & 57 &
30 & 18 & 15 
& 29 & 16 & 13 \\
HOP+ \cite{HOP+}
& 3.49 & - & 67 & 61
& 3.71 & - & 66 & 60 &
36 & 22 & 19
& 34 & 20 & 17 \\
DUET \cite{Map_Memory}
& 3.31 & 81 & 72 & 60
& 3.65 & 76 & 69 & 59 &
46 & 32 & 23
& 52 & 31 & 22 \\
BEVBert \cite{Imagine4}
& 2.81 & 84 & 75 & 64
& 3.13 & 81 & 73 & 62 &
51 & 34 & 24 
& 52 & 32 & 22 \\
Lily \cite{Lily}
& 2.48 & 84 & 77 & 72
& 3.05 & 82 & 74 & 68 &
48 & 32 & 23
& 45 & 30 & 21 \\
ScaleVLN \cite{SCALE}
& 2.34 & 87 & 79 & 70
& 2.73 & 83 & 77 & 68 &
57 & - & -
& 56 & - & - \\
\textbf{SALI (Ours)}
& \textbf{1.92} & \textbf{86} & \textbf{82} & \textbf{78}
& \textbf{2.08} & \textbf{83} & \textbf{79} & \textbf{74} &
\textbf{58} & \textbf{38} & \textbf{28} 
& \textbf{56} & \textbf{34} & \textbf{25} \\
\bottomrule
\end{tabular}
\caption{Comparison with SoTA methods on R2R and REVERIE benchmarks.}
\label{tab:1}
\end{table*}

\subsubsection{3.3.2 Imagination Model Pre-training}
\

\indent \textit{1) Inpaint Model. } With random noise added to the input, we trained the inpaint model by minimizing a combination of cross-entropy loss and mean absolute error (MAE):
\begin{align}
    \mathcal{L}_{inpaint} = -\lambda \sum\nolimits_{i} \hat{s}_i \log(s_t)+(1-\lambda) \Vert d_t-\hat{d}_t\Vert_1,
\end{align}
where $\lambda$ is weight parameter. $s_{t}, d_{t}$ are generated semantic and depth images, while $\hat{s}_{t}, \hat{d}_{t}$ are ground-truth images.

For room-type model training, we get the one-hot codes of room-type for each viewpoint from the Matterport3D Simulator \cite{VLN_R2R}. Two ResNet-50 networks \cite{Resnet}, which have been previously trained on the ImageNet dataset \cite{ImageNet}, are employed to encode the semantic image $s$ and the depth image $d$, respectively. The room-type model takes in the encoded feature and outputs one-hot codes. We calculate cross-entropy loss between the predicted label $l$ and true room type label $\hat{l}$:

\begin{align}
    \mathcal{L}_{room} = -\lambda \sum\nolimits_{i} \hat{l}_i \log(l_i).
\end{align}

\textit{2) Spade Model. } The spade model is a GAN-based model. We train the generator with GAN hinge loss, feature matching loss, and perceptual loss \cite{Predict1}. For PatchGAN-based discriminator \cite{PatchGAN}, we calculate the loss for ground-truth images $\hat{r}_t$ and generated images $r_t$, where $\phi^i$ and $D^i$ denote the output of the $i$-th layer of the pre-trained VGG-19 model and discriminator:
\begin{align}
    \mathcal{L}_{spade}^d = & -\mathbb{E}[\min(0,-1+D(\hat{r}_t))] \nonumber \\
    & - \mathbb{E}[\min(0,-1+D(r_t))], \\
    \mathcal{L}_{spade}^g = & -\lambda_G \mathbb{E}[D(r_t)] \nonumber \\
    & + \lambda_F \sum\nolimits_i^n\frac{\Vert \phi^i(\hat{r}_t)-\phi^i(r_t)\Vert_1}{n} \nonumber \\
    & + \lambda_P \sum\nolimits_i^n\frac{\Vert D^i(\hat{r}_t)-D^i(r_t)\Vert_1}{n}.
\end{align}

We implement the waypoint prediction model following \cite{Bridge}. For each viewpoint, we labeled its neighboring points' relative headings and distances into a 120 $\times$ 12 matrix representing 360 degrees and 3 meters (each element represents 3 degrees and 0.25 meters). A heat map, designated as $\hat{m} \in \mathbb{R}^{120\times 12}$, is then generated as the ground-truth data through interpolation of the aforementioned matrix. For the training process, the above-mentioned two ResNet-50 networks are employed to encode the RGB image and depth image for embedding, $e^r$ and $e^d$. Subsequently, $e^r, e^d$ are merged by a non-linear layer and fed into the model, which generates a heatmap, $m\in \mathbb{R}^{120 \times 12}$. Subsequently, non-maximum suppression (NMS) \cite{NMS} is applied to obtain neighboring waypoints. The loss is calculated by:

\begin{align}
    \mathcal{L}_{waypint} = \frac{1}{120\times12} \sum\nolimits_{i=1}^{120} \sum\nolimits_{j=1}^{12} (m_{ij} - \hat{m}_{ij})^2.
\end{align}

\subsubsection{3.3.3 Inference}
During inference, the agent constructs a global map of an imagination-based memory on-the-fly. It then reasons about the next action over the map, as explained in Section 3.1. As the memory expands, SALI will approach the correct point and find the shortest path to the target. The agent will stop if it selects the `stop' node or reaches the max action steps.

\section{Experiments}
\subsection{Task Setup}
\textbf{Datasets. } We evaluate SALI on VLN benchmarks with both fine-grained instructions (R2R) \cite{VLN_R2R} and coarse ones (REVERIE) \cite{VLN_RVR}. R2R provides step-by-step instructions with an average length of 32 words. REVERIE gives pre-defined object bounding boxes and instructions to guide agents in describing target objects’ positions. Instructions consist of 21 words on average.

\noindent\textbf{Evaluation Metrics. }  We adopt the evaluation metrics \cite{anderson2018metric} used commonly by existing works: 1) Navigation Error (\texttt{NE}) calculates the distance between stop locations and target ones; 2) Trajectory Length (\texttt{TL}) represents the average path length; 3) Success Rate (\texttt{SR}) shows the ratio of stopping within 3m to the target; 4) Success rate weighted by Path Length (\texttt{SPL}) makes trade-off between \texttt{SR} and \texttt{TL}; 5) Oracle Success Rate (\texttt{OSR}) is the ratio of including a viewpoint along the path where the target position can be seen; 6) Remote Grounding Success (\texttt{RGS}) is the ratio of successfully executed instructions; 7) \texttt{RGSPL} is \texttt{RGS} penalized by path length.

\noindent\textbf{Implementation Details. } We utilize ViT-B/16 \cite{ViTB} and ResNet-50 \cite{Res50} to extract node features \( f_t \), and adopt LXMERT \cite{lxmert} as our cross-modal transformer. During pre-training, we set the imagination history length $K$ to 2. SALI is initially trained for 100k iterations with a batch size of 32 using a single Quadro RTX 8000 GPU, alongside training four imagination models. For fine-tuning, the trainable components include imagination node encoders, text encoders, and map encoders. Fine-tuning is conducted for 20k iterations with a batch size of 4 on four Quadro RTX 8000 GPUs. The best model checkpoint is selected based on \(\texttt{SPL} + \texttt{SR}\).

\begin{figure}[!t]
\centering
\includegraphics[width=0.45\textwidth]{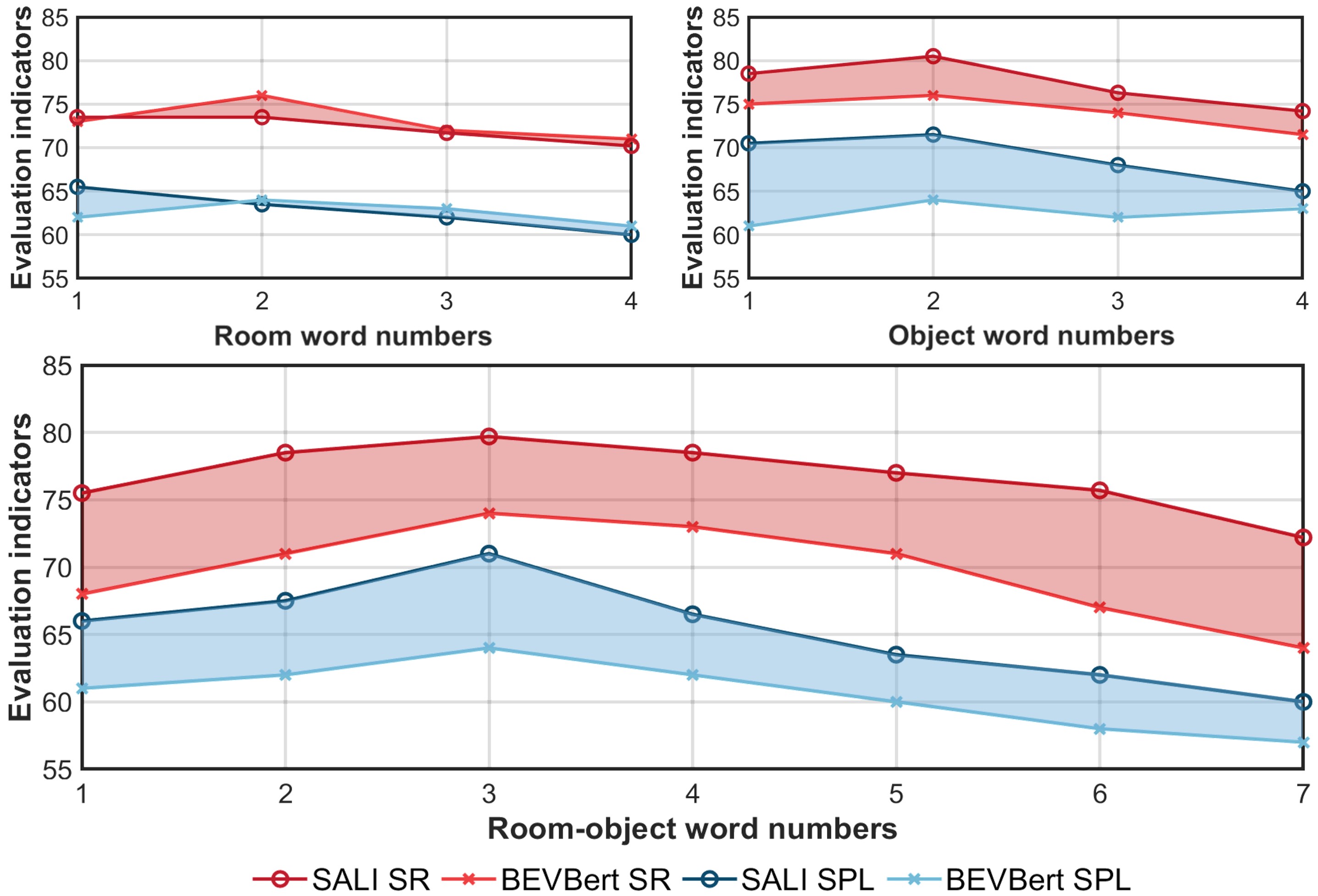} % Reduce the figure size so that it is slightly narrower than the column.
\caption{Comparison of \texttt{SR} and \texttt{SPL} between SALI and BEVBert models on different sub-datasets.}
\label{fig5}
\end{figure}

\subsection{Comparison with State-of-the-Art}
\textbf{R2R.} \Cref{tab:1} compares SALI's navigation performance with various state-of-the-art approaches on the R2R benchmark. SALI achieves superior performance across all metrics. Notably, the concurrent improvement in \texttt{SR} and \texttt{SPL} demonstrates that SALI effectively aligns scenes with instructions, resulting in more efficient navigation.

\noindent\textbf{REVERIE. } SALI also outperforms all previous models on the REVERIE benchmark, achieving the highest scores for \texttt{RGS} and \texttt{RGSPL}. This improvement is attributed to the imagination module, which enhances the agent's ability to recognize and associate objects within its environment.

\subsection{Quantitative and Qualitative Analysis}
\
\indent SALI's episodic memory and simulation capabilities are evidenced by its ability to process complex environment-related commonsense, including associations between rooms, objects, and their spatial relationships.

\noindent\textbf{Quantitative Study. } We examine agents' navigation performance under instructions with complex object and room information. We divided the R2R validation unseen instruction set into:
\begin{itemize}
    \item $\mathcal{S}_1$: Instructions with over two room-related terms and no object terms (e.g. ``\textit{Walk through \textbf{hallway} and towards \textbf{restroom}.}" ).
    \item $\mathcal{S}_2$: Instructions with over two object-related terms and no room terms (e.g. ``\textit{Turn left at the \textbf{oven} and past the \textbf{fridge}.}" ).
    \item $\mathcal{S}_3$: Complex instructions with over four combined room and object terms (e.g. ``\textit{Past the \textbf{door} next to the \textbf{TV} in the \textbf{living room}, then walk into the \textbf{kitchen}.}" ) 
\end{itemize}

Performance on these instruction sets is illustrated in \Cref{fig5}. Compared to the prior best space-aware model (BEVBert, \cite{Imagine4}), SALI achieves the largest performance improvement on $\mathcal{S}_3$ particularly in \texttt{SR} and \texttt{SPL}. SALI’s ability to imagine spatial relationships allows it to effectively interpret intricate object-room associations, demonstrating improved navigation under complex instructions. This showcases SALI’s commonsense reasoning capabilities, achieved through its imagination-memory mechanism, analogous to human episodic simulation memory.
 
\begin{figure*}[!t]
\centering
\includegraphics[width=0.99\textwidth]{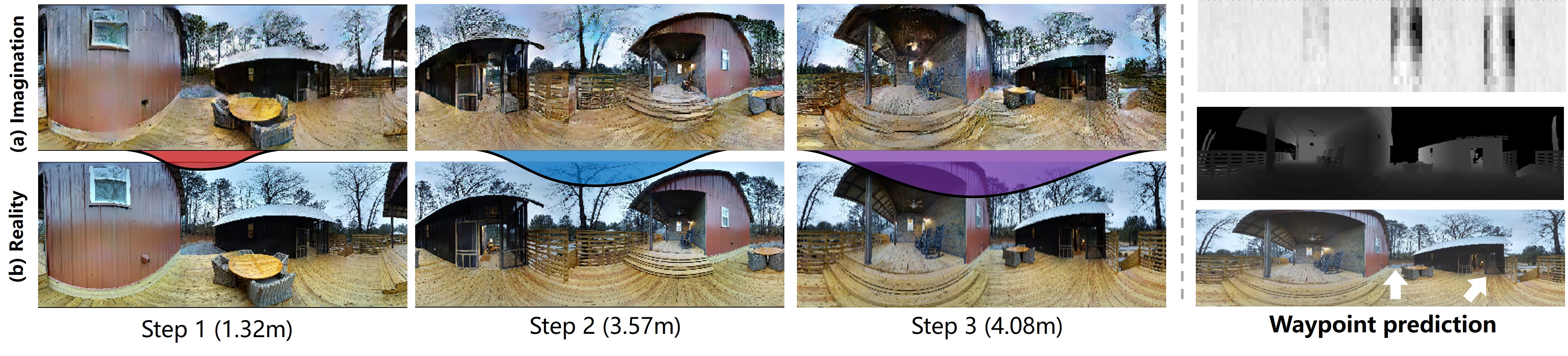} % Reduce the figure size so that it is slightly narrower than the column.
\caption{Imagined picture display and waypoint prediction schematic. The imagination error is visualized by the curve.}
\label{fig6}
\end{figure*}

\noindent\textbf{Qualitative Analysis. }
\Cref{fig6} visualizes the imagination mechanism's process of generating spatial and image information. On the left, the imagined images for goal positions at varying distances are compared with corresponding real images. The peak signal-to-noise ratio (PSNR) \cite{PSNR} of the generated image and the Pearson correlation coefficient \cite{Pearson} with the real image were calculated. The shaded curve region represents the mean and variance of pixel errors of the two images, which can be approximated by visualizing the Pearson correlation coefficient and PSNR. For spatial waypoint prediction, the predicted heat map $m_t$ after the non-maximum suppression (NMS) operation (top right) is visualized to show neighboring navigable positions.

\subsection{Ablation Study}
Extensive experiments on the R2R val-unseen split were conducted to evaluate the proposed features.

\noindent\textbf{Imagination vs. Reality. } We first investigated the role of the hybrid mechanism between imagination and memory in improving navigation performance, as shown in \Cref{tab:2}. It indicates that the memory-based imaginary mechanism gets the SoTA performance, which is 12\% and 9\% higher than ``reality only'' in \texttt{SR} and \texttt{SPL}. At the same time, we find that navigation using only imagination is the least effective. We attribute this to the inability of the agent to effectively use and correlate the imagination results with its experience and trajectory. This suggests it's necessary to establish long-term memory mechanisms supporting imaginative abilities.

\begin{table}[!htbp]
\centering
\fontsize{9pt}{11pt}\selectfont % Set the font size to 9pt
\setlength{\tabcolsep}{1mm} % Set the space between columns
\begin{tabular}{cl | ccccc}
\toprule
\multicolumn{1}{c}{\#} & \multicolumn{1}{c}{Memory Type} &
\multicolumn{1}{c}{TL} & \multicolumn{1}{c}{NE$\downarrow$} & \multicolumn{1}{c}{OSR$\uparrow$} & \multicolumn{1}{c}{\textbf{SR}$\uparrow$} & \multicolumn{1}{c}{\textbf{SPL}$\uparrow$} \\
\midrule
1 & \textbf{Reality} & 12.59 & 3.39 & 78 & 70 & 61 \\
2 & \textbf{Imagination} & 15.71 & 5.44 & 67 & 58 & 50  \\
3 & \textbf{Reality + Imagination} & 10.34 & 1.92 & 86 & 82 & 70  \\
\bottomrule
\end{tabular}
\caption{Ablation study of imagination-based memory.}
\label{tab:2}
\end{table}

\noindent\textbf{Temporal and Spatial Imaginary. } We investigated the impact of the spatial-temporal memory range on navigation performance, where the navigation output time step $M$ and the upper node limit $\bar{N}$ define the memory range (\Cref{tab:3}), fixed input length $K=2$. Results show that performance improves as the memory range expands. However, excessively large ranges ($M=2, \bar{N}=8$) lead to a decline in \texttt{SPL}, as agents at different positions tend to imagine the same destination, interfering with optimal decision-making. While navigation performance must be balanced with training time per epoch, inference time remains stable and does not significantly affect the results.

\begin{table}[!htbp]
\centering
\fontsize{9pt}{11pt}\selectfont
\setlength{\tabcolsep}{1mm}
\begin{tabular}{ccc | ccccc}
\toprule
\multicolumn{1}{c}{\#} & \multicolumn{1}{c}{\textbf{$M$}} & \multicolumn{1}{c}{\textbf{$\bar{N}$}} & 
\multicolumn{1}{c}{OSR$\uparrow$} & \multicolumn{1}{c}{\textbf{SR}$\uparrow$} & \multicolumn{1}{c}{\textbf{SPL}$\uparrow$} & \textbf{Training Time} & \textbf{Inference Time} \\
\midrule
1 & 0 & 0    & 78 & 70 & 61 & 0.54h & 0.15h \\
2 & 1 & 4    & 82 & 76 & 67 & 0.74h & 0.18h \\
3 & 2 & 4   & 86 & 82 & 71 & 1.32h &  0.25h \\
4 & 2 & 8   & 84 & 82 & 68 & 2.51h &  0.30h \\
\bottomrule
\end{tabular}
\caption{The effect of imagination range.}
\label{tab:3}
\end{table}

\noindent\textbf{Auxiliary Model Effectiveness. } The room type model and the waypoint model in the imagination model are auxiliary models that are not necessary for image generation. Waypoint predictions can be made by setting random directions instead. We examined the effect of the two models, as shown in \Cref{tab:4}. We found the agent containing the auxiliary model is optimal in the \texttt{SPL} metrics, reflecting that these two models are essential for decision-making. 

\begin{table}[!htbp]
\centering
\fontsize{9pt}{11pt}\selectfont 
\setlength{\tabcolsep}{1mm} 
\begin{tabular}{cl | ccccc}
\toprule
\# & Model Type & TL & NE$\downarrow$ & OSR$\uparrow$ & \textbf{SR}$\uparrow$ & \textbf{SPL}$\uparrow$ \\
\midrule
1 & \textbf{None} & 12.59 & 3.39 & 78 & 70 & 61 \\
2 & \textbf{Room} & 12.32 & 2.88 & 82 & 75 & 64 \\
3 & \textbf{Waypoint} & 11.85 & 2.54 & 84 & 77 & 66 \\
4 & \textbf{Room + Waypoint} & 10.34 & 1.92 & 86 & 80 & 71 \\
\bottomrule
\end{tabular}
\caption{Ablation study of imagination auxiliary models.}
\label{tab:4}
\end{table}

\noindent\textbf{Dynamic Decision-making. } We evaluate the impact of combining imagined and real node embeddings on navigation performance using static and dynamic weighting strategies. Results show that dynamic weighting is more effective, as the weight of the imagination module decreases over time, reflecting reduced reliance on imagination in later navigation stages, analogous to human memory mechanisms.

\begin{table}[!htbp]
\centering
\fontsize{9pt}{11pt}\selectfont 
\setlength{\tabcolsep}{1mm} 
\begin{tabular}{cl | ccccc}
\toprule
\# & Decisioning Weight &
TL & NE$\downarrow$ & OSR$\uparrow$ & \textbf{SR}$\uparrow$ & \textbf{SPL}$\uparrow$ \\
\midrule
1 & \textbf{Dynamic} & 10.34 & 1.92 & 86 & 80 & 71 \\
2 & \textbf{Fixed ($\gamma_t=0.5$)} & 12.33 & 2.25 & 82 & 76 & 67 \\
\bottomrule
\end{tabular}
\caption{Ablation study of Decision Weight.}
\label{tab:5}
\end{table}

\section{Conclusion} We propose SALI, an agent equipped with an imagination-integrated memory mechanism, inspired by human episodic memory and episodic simulation. Extensive experimental results demonstrate the effectiveness of the imagination module, enabling SALI to achieve SoTA performance in unseen environments. By leveraging its hybrid memory, SALI enhances robustness when navigating in complex environments. The future work will focus on optimizing the computational efficiency of the imagination process while preserving fine-grained results. 

\section*{Acknowledgments}
This work was supported in part by the Natural Science Foundation of China under Grant 62303307, in part by National Key R\&D Program of China under Grant No.2023YFB4705700, in part by Shanghai Municipal Science and Technology Major Project under Grant 2021SHZDZX0102 and in part by the Fundamental Research Funds for the Central Universities.

\bibliography{aaai25}

\end{document}